\UseRawInputEncoding
\def\year{2021}\relax
\documentclass[letterpaper]{article}
\usepackage{aaai21} 
\usepackage{times} 
\usepackage{helvet} 
\usepackage{courier} 
\usepackage[hyphens]{url} 
\usepackage{graphicx} 
\usepackage{graphicx} 
\usepackage{natbib}
\usepackage{caption}
\usepackage{booktabs}       
\usepackage{amsfonts}       
\usepackage{nicefrac}       
\usepackage{microtype}      
\usepackage{amsmath}

\DeclareMathOperator*{\argmin}{arg\,min}

\title{Sparsely constrained neural networks for model discovery of PDEs}

\author {
    Gert-Jan Both,\textsuperscript{\rm 1}
    Gijs Vermari\"en, \textsuperscript{\rm 2}
    Remy Kusters \textsuperscript{\rm 1} \\
}
\affiliations {
    \textsuperscript{\rm 1} Université de Paris, INSERM U1284, Center for Research and Interdisciplinarity (CRI), F-75006 Paris, France \\
    \textsuperscript{\rm 2} Leiden Observatory, Leiden University, Leiden, The Netherlands \\
    gert-jan.both@cri-paris.org, vermarien@strw.leidenuniv.nl, remy.kusters@cri-paris.org
}
\begin{document}

\maketitle

\begin{abstract}

Sparse regression on a library of candidate features has developed as the prime method to discover the partial differential equation underlying a spatio-temporal data-set. These features consist of higher order derivatives, limiting model discovery to densely sampled data-sets with low noise. Neural network-based approaches circumvent this limit by constructing a surrogate model of the data, but have to date ignored advances in sparse regression algorithms. In this paper we present a modular framework that dynamically determines the sparsity pattern of a deep-learning based surrogate using any sparse regression technique. Using our new approach, we introduce a new constraint on the neural network and show how a different network architecture and sparsity estimator improve model discovery accuracy and convergence on several benchmark examples. Our framework is available at \url{https://github.com/PhIMaL/DeePyMoD}
\end{abstract}

\section{Introduction}

Model discovery aims at finding interpretive models in the form of PDEs from large spatio-temporal data-sets. Most algorithms apply sparse regression on a predefined set of candidate terms, as initially proposed by Brunton et al. for ODEs with SINDY \cite{brunton_discovering_2016} and by Rudy et al. for PDEs with PDE-find \cite{rudy_data-driven_2017}. By writing the unknown differential equation as $\partial_t u = f(u, u_x, ...)$ and assuming the right-hand side is a linear combination of predefined terms, i.e. $f(u, u_x, ...) = au + bu_x + ... =  \Theta \xi$, model discovery reduces to finding a sparse coefficient vector $\xi$. Calculating the time derivative $u_t$ and the function library $\Theta$ is notoriously hard for noisy and sparse data since it involves calculating higher order derivatives. The error in these terms is typically high due to the use of numerical differentiation techniques such as finite difference or spline interpolation, limiting classical model discovery to low-noise and densely sampled data-sets. Deep learning-based methods circumvent this issue by constructing a surrogate from the data and calculating the feature library $\Theta$ as well as the time derivative $u_t$ from this digital twin using automatic differentiation. This approach significantly improves the accuracy of the time derivative and the library in noisy and sparse data sets, but suffers from convergence issues and, to date, does not leverage advanced sparse regression techniques.

\begin{figure*}
    \centering
    \includegraphics[width=0.9 \textwidth]{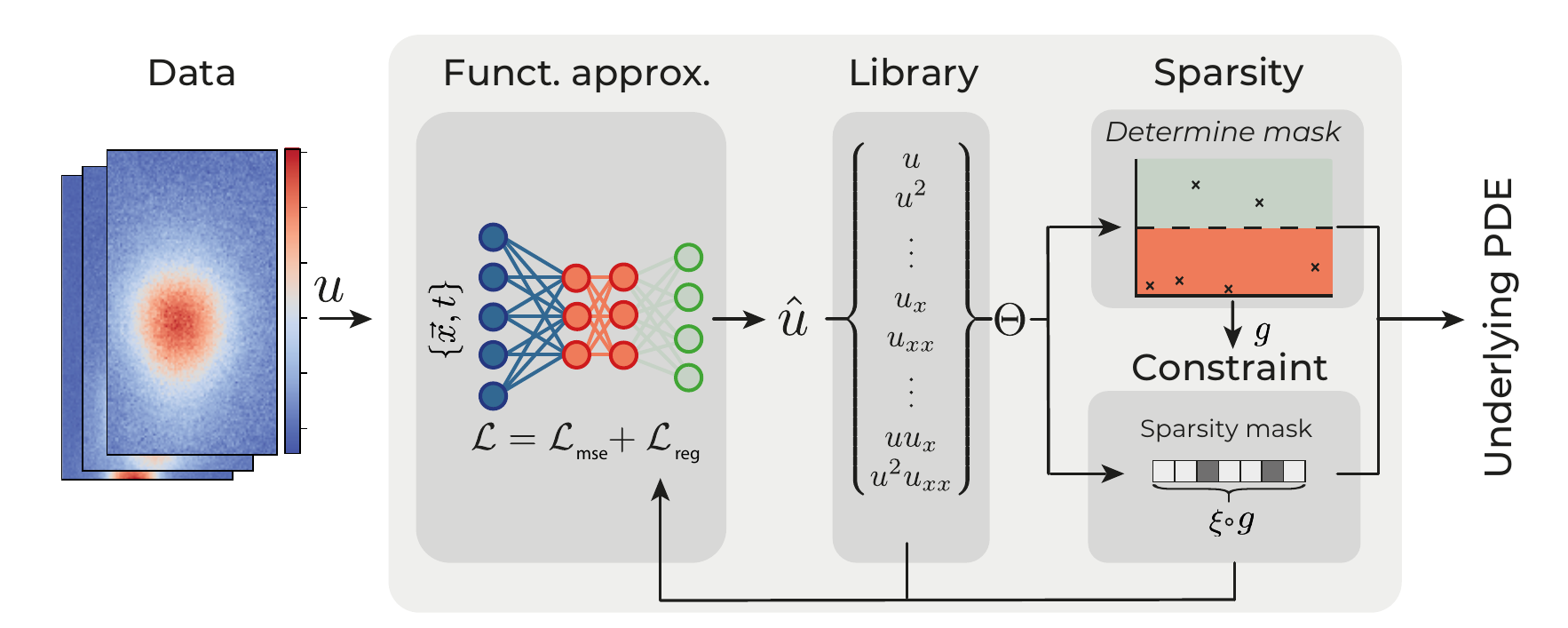}
    \caption{Schematic overview of our framework. (I) A \textbf{function approximator} constructs a surrogate of the data, (II) from which a \textbf{Library} of possible terms and the time derivative is constructed using automatic differentiation. (III) A \textbf{sparsity estimator} selects the active terms in the library using sparse regression and (IV) the function approximator is constrained to solutions allowed by the active terms by the  \textbf{constraint}.}
    \label{fig:schematic}
\end{figure*}

In this paper we present a modular approach to combine deep-learning based models with state-of-the-art sparse regression techniques. Our framework consists of a neural network to model the data, from which we construct the function library. Key to our approach is that we dynamically apply a mask to select the active terms in the function library throughout training and constrain the network to solutions of the equation given by these active terms. To determine this mask, we can use any non-differentiable sparsity-promoting algorithm (see figure \ref{fig:schematic}). This allows us to use a constrained neural network to model the data and construct an accurate function library, while an advanced sparsity promoting algorithm is used to dynamically discover the equation based on output from the network.

We present three experiments to show how varying these components improves the performance of model discovery. (I) We replace the gradient-based optimisation of the constraint by one based on ordinary least squares, leading to much faster convergence. (II) We show that using PDE-find to find the active components outperforms a threshold-based Lasso approach in highly noisy data-set. (III) We demonstrate that using a SIREN \cite{sitzmann_implicit_2020} instead of a standard feed forward-neural network allows us to discover equations from highly complex data-sets.

\subsection{Related Work}

\paragraph{Sparse regression}

Sparse regression as a means to discover differential equations was pioneered by SINDY \cite{brunton_discovering_2016} and PDE-find \cite{rudy_data-driven_2017}. They have since been expanded to automated hyper-parameter tuning \cite{champion_data-driven_2019, maddu_stability_2019}; a Bayesian approach for model discovery using Sparse Bayesian Learning \cite{yuan_machine_2019}, model discovery for parametric differential equations\cite{rudy_deep_2019} and evolutionary  approach to PDE discovery \cite{maslyaev_data-driven_2019}. 

\paragraph{Deep learning-based model discovery} With the advent of Physics Informed neural networks \cite{raissi_physics_2017, raissi_physics_2017-1}, a neural network has become one of the prime approaches to create a surrogate of the data and then perform sparse regression on the networks prediction \cite{schaeffer_learning_2017, berg_data-driven_2019}. Alternatively, Neural ODEs are introduced to discover unknown governing equation \cite{rackauckas_universal_2020} from physical data-sets. Different optimisation strategy based on the method of alternating direction is considered in \cite{chen_deep_2020}, and graph based approaches have been developed recently \cite{seo_differentiable_2019, sanchez-gonzalez_graph_2018}. \cite{greydanus_hamiltonian_2019} and \cite{cranmer_lagrangian_2020} directly encode symmetries in neural networks using respectively the Hamiltonian and Lagrangian framework. Finally, auto-encoders have been used to model PDEs and discover latent variables\cite{lu_extracting_2019, iten_discovering_2020}, but do not lead to an explicit equation and require large amounts of data.

\section{Deep-learning based model discovery with sparse regression}

Deep learning-based model discovery uses a neural network to construct a surrogate model $\hat{u}$ of the data $u$. A library of candidate terms $\Theta$ is constructed using automatic differentiation from $\hat{u}$ and the neural network is constrained to solutions allowed by this library \cite{both_deepmod:_2019}. The loss function of the network thus consists of two contributions, (i) a mean square error to learn the mapping $(\vec{x},t) \rightarrow \hat{u}$ and (ii) a term to constrain the network,

\begin{equation}
\mathcal{L} = \frac{1}{N}\sum_{i=1}^{N}\left( u_i - \hat{u}_i \right) ^2 +\frac{1}{N}\sum_{i=1}^{N}\left( \partial_t \hat{u}_i - \Theta_{i}\xi \right)^2 .
 \label{eq:deepmod}
\end{equation}

 The sparse coefficient vector $\xi$ is learned concurrently with the network parameters and plays two roles: 1) determining the active (i.e. non-zero) components of the underlying PDE and 2) constraining the network according to these active terms. We propose to separate these two tasks by \emph{decoupling the constraint from the sparsity selection process itself}. We first calculate a sparsity mask $g$ and constrain the network only by the active terms in the mask: instead of constraining the neural network with $\xi$, we constrain it with $\xi \circ \ g$, replacing eq. \ref{eq:deepmod} with
 
 \begin{equation}
\mathcal{L} = \frac{1}{N}\sum_{i=1}^{N}\left( u_i - \hat{u}_i \right) ^2 +\frac{1}{N}\sum_{i=1}^{N}\left( \partial_t \hat{u}_i - \Theta_{i}(\xi \cdot g)\right)^2 .
 \label{eq:deepmod2}
\end{equation}

Training using eq. \ref{eq:deepmod2} requires two steps: first, we calculate $g$ using a sparse estimator. Next, we minimise it with respect to the network parameters using the masked coefficient vector. The sparsity mask $g$ need not be calculated differentiably, so that any classical, non-differentiable sparse estimator can be used. This approach has several additional advantages: i) It provides an unbiased estimate of the coefficient vector since we do not apply $l_1$ or $l_2$ regularisation on $\xi$, ii) the sparsity pattern is determined from the full library $\Theta$, rather than only from the remaining active terms, allowing dynamic addition and removal of active terms throughout training, and iii) we can use cross validation in the sparse estimator to find the optimal hyper parameters for model selection. Finally, we note that the sparsity mask $g$ mirrors the role of attention in transformers \cite{bahdanau_neural_2016}.

Using this change, we construct a general framework for deep learning based model discovery using four modules (see figure \ref{fig:schematic}). (I) A \textbf{function approximator} constructs a surrogate model of the data, (II) from which a \textbf{Library} of possible terms and the time derivative is constructed using automatic differentiation. (III) A \textbf{sparsity estimator} constructs a sparsity mask to select the active terms in the library using some sparse regression algorithm and (IV) a \textbf{constraint} constrains the function approximator to solutions allowed by the active terms obtained from the sparsity estimator.

\paragraph{Training} We typically calculate the sparsity mask $g$ using an external, non-differentiable estimator. In this case, updating the mask at the right time is crucial: before the function approximator has reasonably approximated the data, updating the mask would adversely affect training, as it is likely to select the wrong terms. Vice versa, updating the mask too late risks using a function library from an overfitted network. We implement a procedure in the spirit of "early stopping" to decide when to update: the data-set gets split into a train and test-set and we update the mask once the mean squared error on the test-set reaches a minimum or changes less than a preset value $\delta$. We typically set $\delta = 10^{-6}$ to ensure the network has learned a good representation of the data.

\begin{figure}[t]
  \begin{center}
    \includegraphics[width=0.45\textwidth]{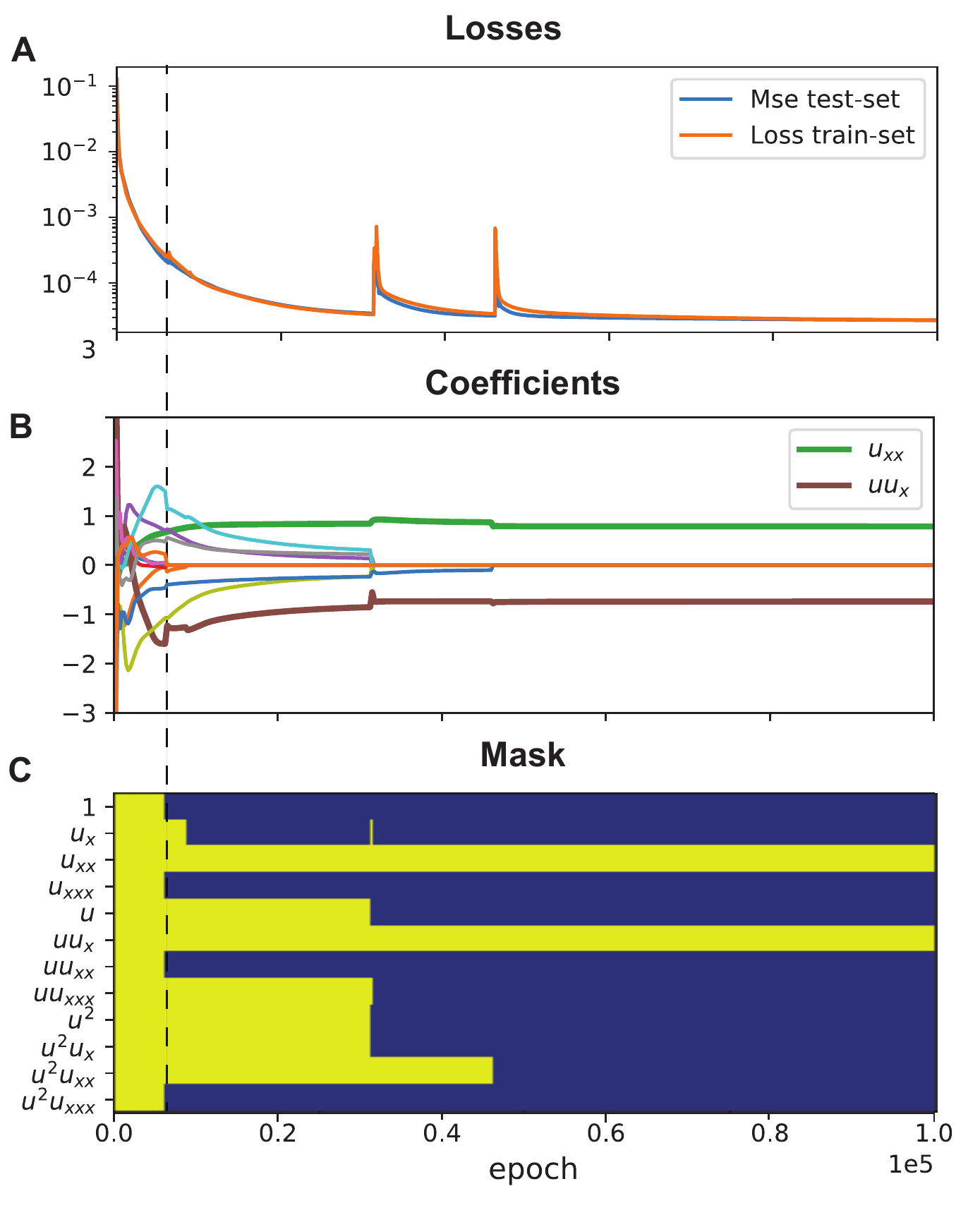}
  \end{center}
  \caption{\textbf{A)} MSE of the test-set and the total loss of the train-set as function of the number of epochs. The vertical line indicates the first time the sparsity mask is applied. \textbf{B)} The twelve coefficients as function of the number of epochs. The two terms $u_{xx}$ and $uu_x$ need to be recovered. \textbf{C)} Dynamic sparsity mask during training. Yellow components are active, blue components are inactive.}
  \label{fig:training}
\end{figure}

After the first update, we periodically update the mask using the sparsity estimator. In figure \ref{fig:training} we demonstrate this training procedure on a Burgers equation with 1500 samples with 2$\%$ white noise. It shows the losses on the train- and testset in panel A, the coefficients of the constraint in panel B and the sparsity mask in C. In practice we observe that large data-sets with little noise typically discover the correct PDE after only a single sparsity update, but that noisy data-sets require several updates, removing only a few terms at a time. Final convergence is reached when the $l_1$ norm of the coefficient vector remains constant.

\paragraph{Package}
We provide our framework as a python based package at \url{https://github.com/PhIMaL/DeePyMoD}, with the documentation and examples available at \url{https://phimal.github.io/DeePyMoD/}. Mirroring our approach, each model consists of four modules: a function approximator, library, constraint and sparsity estimator module. Each module can be customised or replaced without affecting the other modules, allowing for quick experimentation. Our framework is built on Pytorch \cite{paszke_pytorch_2019} and any Pytorch model (i.e. Recurrent Neural Networks) can be used as function approximator. The sparse estimator module follows the Scikit-learn API \cite{pedregosa_scikit-learn_nodate, buitinck_api_2013}, i.e., all the build-in Scikit-learn estimators, such as those in PySindy\cite{de_silva_pysindy_2020} or SK-time \cite{loning_sktime_nodate}, can be used. 

\section{Experiments}

\begin{figure}
  \begin{center}
    \includegraphics[width=0.35\textwidth]{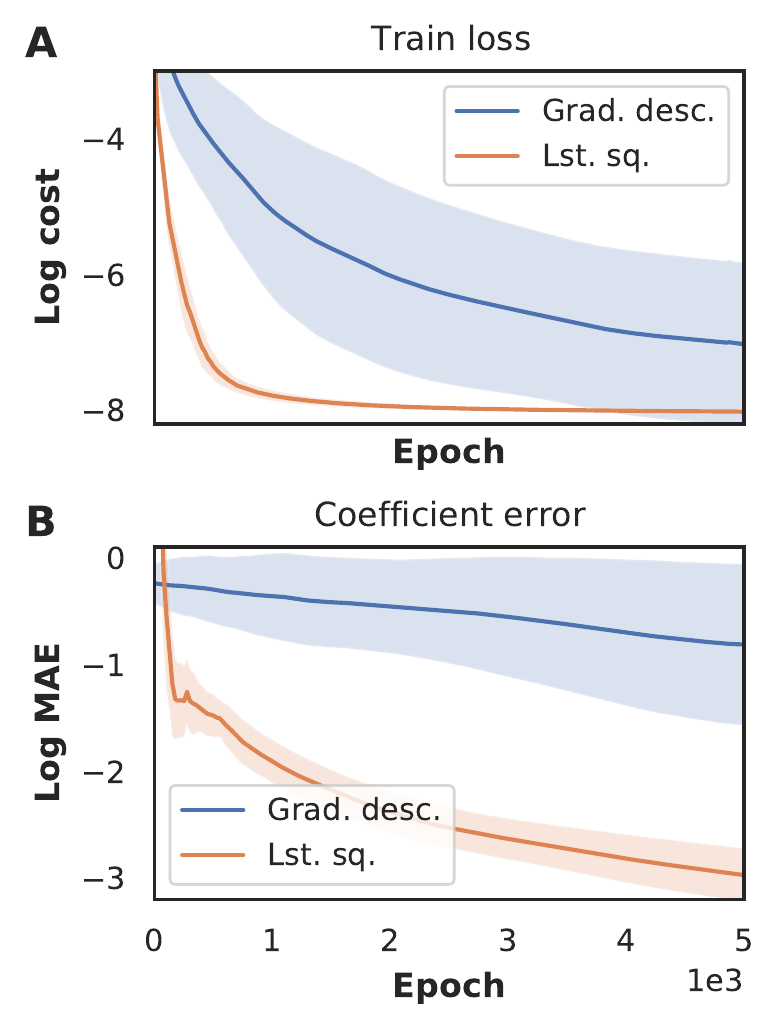}
  \end{center}
  \caption{\textbf{A)} Loss and \textbf{B)} mean absolute error of the coefficients obtained with the gradient descent and the least squares constraint as a function of the number of epochs. Results have been averaged over twenty runs and shaded area denotes the standard deviation.}
  \label{fig:lstsq_vs_grad}
\end{figure}

\paragraph{Constraint}
The sparse coefficient vector $\xi$ in eq. \ref{eq:deepmod} is typically found by optimising it concurrently with the neural network parameters $\theta$. Considering a network with parameter configuration $\theta^*$, the problem of finding $\xi$ can be rewritten as $\argmin_{\xi} \left|u_t(\theta^*) - \Theta (\theta^*)\xi\right|^2$. This can be analytically solved by least squares under mild assumptions; we calculate $\xi$ by solving this problem every iteration, rather than optimizing it using gradient descent. In figure \ref{fig:lstsq_vs_grad} we compare the two constraining strategies on a Burgers data-set\footnote{We solve $u_t = u_xx + \nu uu_x$ with a delta-peak initial condition for $\nu=0.1$ for $x =[-3, 4]$, $t=[0.5, 5]$, randomly sample 2000 points and add $10\%$ white noise.}, by training for 5000 epochs without updating the sparsity mask\footnote{All experiments use a network with a $\tanh$ activation function of 5 layers with 30 neurons per layer. The network is optimized using the ADAM optimiser with a learning rate of $2\cdot10^{-3}$ and $\beta = (0.99,0.999)$.}. Panel A) shows that the least-squares approach reaches a consistently lower loss. More strikingly, we show in panel B) that the mean absolute error in the coefficients is three orders of magnitude lower. We explain the difference as a consequence of the random initialisation of $\xi$: the network is initially constrained by incorrect coefficients, prolonging convergence. The random initialisation also causes the larger spread in results compared to the least squares method. The least squares method does not suffer from sensitivity to the initialisation and consistently converges.

\paragraph{Sparsity estimator} 

\begin{figure}
  \begin{center}
    \includegraphics[width=0.35\textwidth]{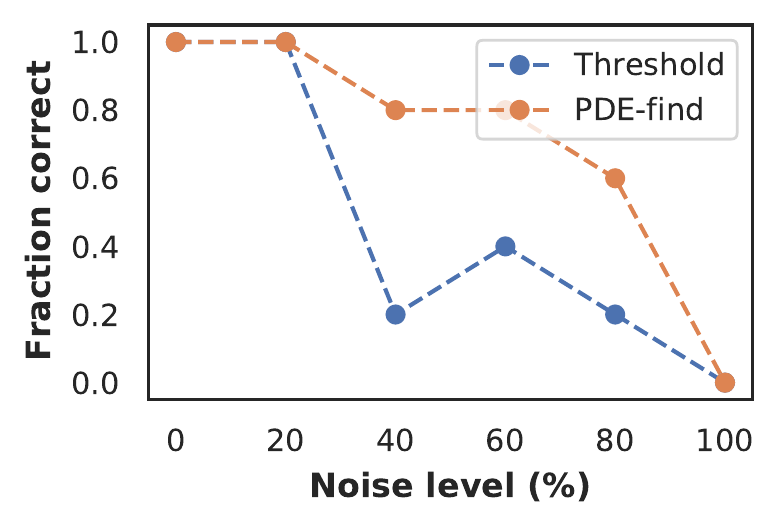}
  \end{center}
  \caption{Fraction of correct discovered Burgers equations (averaged over 10 runs) as function of the noise level for the thresholded lasso and PDE-find sparsity estimator.}
  \label{fig:Burgers}
\end{figure}

Implementing the sparsity estimator separately from the neural network allows us to use any sparsity promoting algorithm. Here we show that a classical method for PDE model discovery, PDE-find \cite{rudy_data-driven_2017}, can be used together with neural networks to perform model discovery in highly sparse and noisy data-sets. We compare it with the thresholded Lasso\footnote{We use a pre-set threshold of 0.1.} in figure \ref{fig:Burgers} approach \cite{both_deepmod:_2019} on a Burgers data-set \footnote{See footnote 2, only with 1000 points randomly sampled.} with varying amounts of noise. The PDE-find estimator discovers the correct equation in the majority of cases, even with up to $60\% - 80\%$ noise, whereas the thresholded lasso mostly fails at $40\%$. We emphasise that the modular approach we propose here allows to combine classical and deep learning-based techniques. More advanced sparsity estimators such as SR3 \cite{champion_unified_2019} can easily be included in this framework. 

\begin{figure*}
    \centering
    \includegraphics[width= \textwidth]{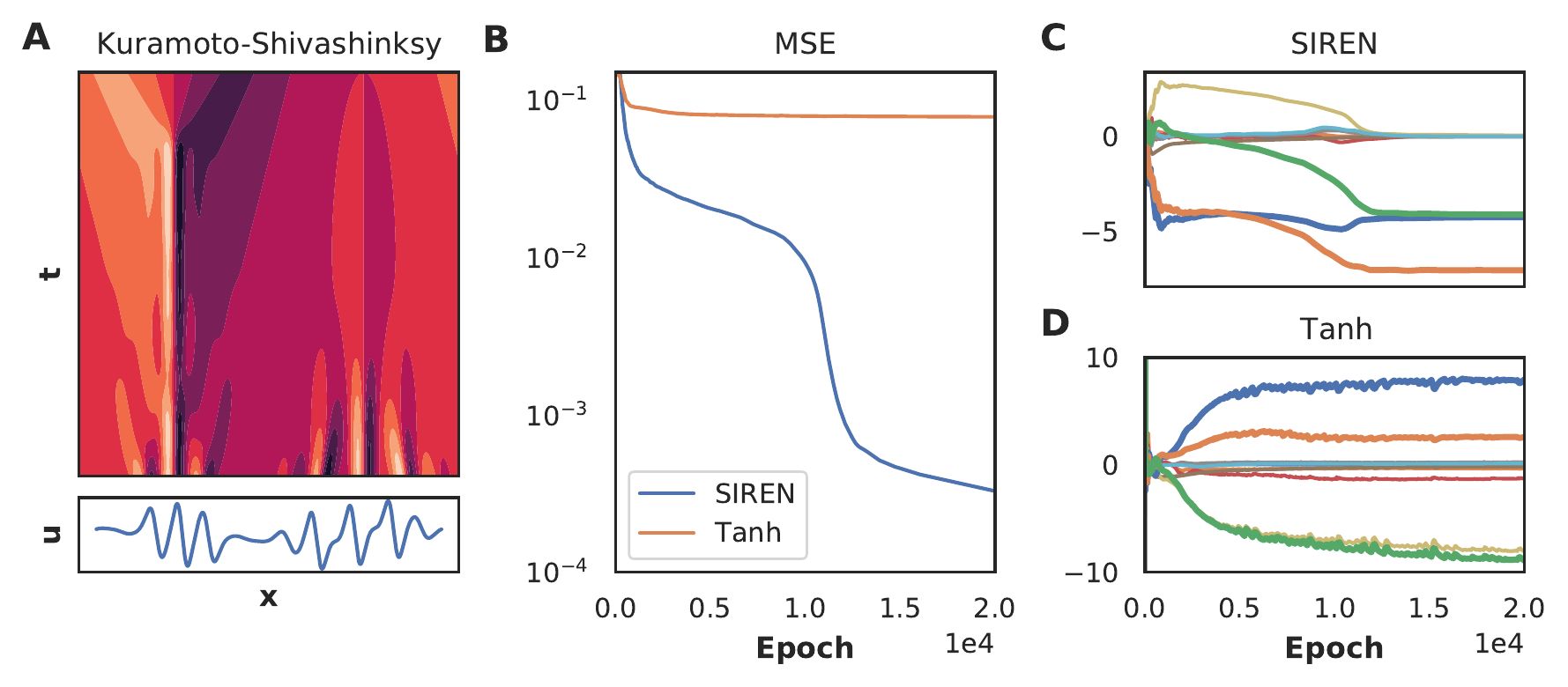}
    \caption{ \textbf{A)} Solution of the KS equation. Lower panel shows the cross section at the last time point: $t=44$. \textbf{B)} MSE as function of the number of epochs for both the tanh-based and SIREN NN. Coefficients as function of number of epochs for \textbf{C)} the SIREN. and \textbf{D)} the tanh-based NN. The bold curves in panel C and D are the terms in the KS equation components; green: $u u_x$:, blue: $u_{xx}$ and orange: $u_{xxxx}$. Only SIREN is able to discover the correct equation.}
    \label{fig:KS}
\end{figure*}

\paragraph{Function approximator} We show in figure \ref{fig:KS} that a $\tanh$-based NN fails to converge on a data-set of the Kuramoto-Shivashinksy (KS) equation\footnote{We solve $\partial_t u + u u_x + u_{xx} + u_{xxxx} = 0 $ between $x=[0, 100], t=[0, 44]$, randomly sample 25000 points and add $5\%$ white noise.}(panel A and B). Consequently, the coefficient vectors are incorrect (Panel D). As our framework is agnostic to the underlying function approximator, we instead use a SIREN \footnote{Both networks use 8 layers with 50 neurons. We train the SIREN using ADAM with a learning rate of $2.5\cdot10^{-4}$ and $\beta=(0.999, 0.999)$}, which is able to learn very sharp features in the underlying dynamics. In panel B we show that a SIREN is able to learn the complex dynamics of the KS equation and in panel C that it discovers the correct equation\footnote{In bold; $u u_x$: green, $u_{xx}$: blue and $u_{xxxx}$: orange}.  This example shows that the choice of function approximator can be a decisive factor in the success of neural network based model discovery. Using our framework we can also explore using RNNs, Neural ODEs \cite{rackauckas_universal_2020} or Graph Neural Networks \cite{seo_differentiable_2019}.

\section{Discussion and future work}
In this paper we introduced a framework for model discovery, combining classical sparsity estimation with deep learning based surrogates. Building on this, we showed that replacing the function approximator, constraint or dynamically applying the sparsity estimator during training can extend model discovery to more complex datasets, speed up convergence or make it more robust to noise. Each of the four components is decoupled from the rest and can be independently changed, making our approach a solid base for future research. Currently, the function approximator simply learns the solution using a feed forward neural network. We suspect that adding more structure, for example by using recurrent, convolutional or graph neural networks, will improve the performance of model discovery. It might also be beneficial to regularise the constraint, for example by implementing lasso or ridge regression. Updating the sparsity mask in a non-differentiable manner works because the neural network is able to learn a fairly accurate surrogate without imposing sparsity on the constraint. If the network is unable to learn an accurate representation, our approach breaks down. Updating the mask in a differentiable manner would not suffer from this drawback, and we intend to pursue this in future works.

\section*{Acknowledgments}
This work received support from the CRI Research Fellowship to attributed to Remy Kusters. We thank the Bettencourt Schueller Foundation long term partnership and NVidia for supplying the GPU under the Academic Grant program. We would also like to thank the authors and contributors of Numpy (\cite{harris_array_2020}), Scipy (\cite{virtanen_scipy_2020}), Scikit-learn (\cite{pedregosa_scikit-learn_nodate}), Matplotlib (\cite{hunter_matplotlib_2007}), Ipython (\cite{perez_ipython_2007}), and Pytorch (\cite{paszke_pytorch_2019}) for making our work possible through their open-source software. The authors declare no competing interest.

\begin{quote}
\begin{small}
\bibliography{references}
\end{small}
\end{quote}

\end{document}